\documentclass[english]{llncs}
\usepackage[T1]{fontenc}
\usepackage[latin9]{inputenc}
\usepackage{array}
\usepackage{float}
\usepackage{rotfloat}
\usepackage{multirow}
\usepackage{amstext}
\usepackage{graphicx}

\makeatletter

\providecommand{\tabularnewline}{\\}
\floatstyle{ruled}
\newfloat{algorithm}{tbp}{loa}
\providecommand{\algorithmname}{Algorithm}
\floatname{algorithm}{\protect\algorithmname}

 \usepackage{algolyx}

\PassOptionsToPackage{obeyspaces,spaces}{url}
\usepackage[obeyspaces,spaces]{url}
\algoption{noend}
\keycomment{/*}{*/}
\usepackage{multirow}
\usepackage{caption}

\makeatother

\usepackage{babel}
\begin{document}
\algsetup{indent=2em}
\renewcommand{\algorithmiccomment}[1]{/*#1*/}

\title{Efficient Partial Order CDCL Using Assertion Level Choice Heuristics\thanks{We gratefully acknowledge the financial support of the Natural Sciences and Engineering Research Council of
Canada on this research.}}

\author{Anthony Monnet \and Roger Villemaire}

\institute{Université du Québec à Montréal, Montreal, Canada\\\email{anthonymonnet@aol.fr}\\\email{villemaire.roger@uqam.ca}}
\maketitle
\begin{abstract}
We previously designed Partial Order Conflict Driven Clause Learning
(PO-CDCL), a variation of the satisfiability solving CDCL algorithm
with a partial order on decision levels, and showed that it can speed
up the solving on problems with a high independence between decision
levels. In this paper, we more thoroughly analyze the reasons of the
efficiency of PO-CDCL. Of particular importance is that the partial
order introduces several candidates for the assertion level. By evaluating
different heuristics for this choice, we show that the assertion level
selection has an important impact on solving and that a carefully
designed heuristic can significantly improve performances on relevant
benchmarks.
\end{abstract}

\section{Introduction}

The SAT problem consists in deciding if a given propositional formula
expressed in conjunctive normal form is satisfiable, i.e. if there
exists a truth assignment that makes the formula true. Furthermore,
a satisfying assignment, or model, has to be returned if the formula
is satisfiable. Many decision problems can be encoded using a propositional
formula, such that this formula is satisfiable iff the considered
problem has a solution. 

Conflict-driven clause learning (CDCL)~\cite{JPMSKAS99} is the algorithm
used by the most efficient complete SAT solvers. Unlike basic depth-first
search that only undoes the last decision when a conflict is reached,
CDCL is able to analyze the reasons for this conflict and to define
the assertion level, which is the second to last decision level involved
in this conflict. It then backtracks directly to this assertion level,
often undoing several decision levels at once, in order to ensure
that this conflict will not be encountered again in this branch of
the search. It therefore performs a much more efficient pruning of
the search space than regular depth-first search, often leading to
a significantly faster solving of problems.

CDCL has however the negative side-effect of destroying parts of the
current partial assignment not directly related to the conflict. Indeed,
by returning straight to the assertion level, it entirely destroys
all instantiations in subsequent decision levels. By definition, none
of them were directly involved in the conflict, except for the decision
level where the conflict was discovered. In the worst case, these
instantiations may even belong to a different connected component
of the problem and couldn't possibly be affected by the conflict resolution,
even indirectly. CDCL thus can cause the unnecessary deletion of previous
parts of the search, which may prevent the detection of some conflicts
or the completion of a satisfying assignment, ultimately slowing down
the solving process.

This deletion of unrelated parts of the search is caused by the implicit
total ordering on successive decisions during the search, and this
total order can be relaxed without damaging the correctness, completeness
and termination of the algorithm. We therefore designed partial order
CDCL (PO-CDCL)~\cite{AMRV12}, a variant of CDCL maintaining a partial
order between decision levels, which allows to locally undo less instantiations
during a conflict-directed backtrack. In practice, some SAT problems
(for instance encodings from the formal verification of superscalar
microprocessors~\cite{MNVREB03}) have a relatively sparse dependency
between decision levels during solving, and we showed that PO-CDCL
significantly decreases the solving time on these instances.

The aim of this paper is twofold. First, we show that the efficiency
of PO-CDCL is due to the fact that it dramatically reduces the search
efforts needed to reach successive conflicts and hence prune the search
space. Secondly, we consider a new parameter of the algorithm introduced
by the partial order: unlike in a regular CDCL, the assertion level
of a conflict clause is not uniquely defined and can be chosen using
various heuristics. We show that this choice has a significant impact
on the search, and that heuristics affecting the average amount of
instantiations undone by conflicts can further significantly improve
the performance of PO-CDCL. Interestingly, the solving of satisfiable
problems is improved when this average amount of undone instantiations
increases, while unsatisfiability is proved faster when it decreases.
Moreover, this quantity is most efficiently controlled indirectly
by choosing assertion levels that maximize or minimize the number
of additional dependencies they would introduce between decision levels.

The remainder of this paper is organized as follows: Section~\ref{sec:PO-CDCL}
introduces the PO-CDCL algorithm. Section~\ref{sec:Related-Works}
reviews related methods seeking to reduce the amount of instantiations
deleted during a non-chronological backtracking algorithm in CSP and
SAT. Finally, section~\ref{sec:Assertion-level-heuristics} presents
experimental results obtained with the implementation of PO-CDCL in
a state-of-the-art CDCL solver using various heuristics for the choice
of the assertion level. These results are used to analyze the causes
of the efficiency of PO-CDCL with various assertion level heuristics
on satisfiable and unsatisfiable SAT instances with low level interdependencies.

\section{PO-CDCL\label{sec:PO-CDCL}}

\begin{algorithm}[p]
\begin{algor}[1]
\item [{{*}}] $\sigma\leftarrow\emptyset$ /* begin with the empty assignment
*/
\item [{{*}}] $\lambda=0$ /* $\lambda$ is the current decision level
*/
\item [{loop}]~

\begin{algor}[1]
\item [{{*}}] \textsc{$c\leftarrow$Propagate /* }propagate new instantiations\textsc{
*/}
\item [{if}] $c\neq\mathrm{NIL}$ /* a conflict was found during propagations
*/

\begin{algor}[1]
\item [{if}] $\text{\ensuremath{\lambda}}=0$ /* conflict at decision
level 0 */

\begin{algor}[1]
\item [{{*}}] \textbf{return} false /* unsatisfiable problem */
\end{algor}
\item [{else}]~

\begin{algor}[1]
\item [{{*}}] $\gamma\leftarrow\textsc{Analyze}(c)$ /* infer the conflict
clause $\gamma$ */
\item [{{*}}] $\mathrm{candidates}\leftarrow$ all $\Delta$-maximal elements
in $\mathrm{levels}(\gamma)\setminus\{\lambda\}$
\item [{{*}}] \textbf{choose} $a$ \textbf{in} candidates /* $a$ is the
assertion level */
\item [{for}] $l>_{\Delta}a$

\begin{algor}[1]
\item [{{*}}] delete level $l$
\end{algor}
\item [{endfor}]~
\item [{{*}}] $\lambda\leftarrow a$ /* $a$ becomes the current level
*/
\item [{{*}}] $\textsc{Learn}(\gamma)$
\item [{{*}}] \textsc{PropagateAssertion}($\gamma)$
\end{algor}
\item [{endif}]~
\end{algor}
\item [{else}] /* no conflict during propagations */

\begin{algor}[1]
\item [{if}] all variables are instantiated

\begin{algor}[1]
\item [{{*}}] \textbf{return} $\sigma$ /* $\sigma$ is a model */
\end{algor}
\item [{else}]~

\begin{algor}[1]
\item [{{*}}] $\lambda\leftarrow\textsc{NewLevel}$
\item [{{*}}] $\textsc{Decide}(\lambda)$
\end{algor}
\item [{endif}]~
\end{algor}
\item [{endif}]~
\end{algor}
\item [{endloop}]~
\end{algor}
\caption{PO-CDCL}
\label{PO-CDCL}
\end{algorithm}

CDCL~\cite{JPMSKAS99} is a satisfiability solving algorithm based
on the older depth-first search DPLL~\cite{MDGLDWL62}, enhanced
with conflict-directed backtracking and clause learning. It successively
assigns arbitrary values to variables (it takes decisions) until either
a clause is violated or all variables are assigned. After each decision,
an exhaustive round of unit propagation is performed to deduce all
possible consequences of the current assignment using this inference
rule. A decision level is the set formed by a decision and all the
propagations it entails.

\begin{algorithm}[p]
\begin{algor}[1]
\item [{{*}}] $\Pi\leftarrow$\{instantiations not yet propagated\}
\item [{while}] \{$\Pi\neq\emptyset$\}

\begin{algor}[1]
\item [{{*}}] choose $l\in\Pi$
\item [{for}] all clauses $c$ s.t. $\neg l$ is watched in $c$

\begin{algor}[1]
\item [{{*}}] $w\leftarrow$ the second watched literal in $c$
\item [{if}] $\sigma(w)=\mbox{true}$

\begin{algor}[1]
\item [{{*}}] \textbf{set} $\mathrm{level}(w)<_{\Delta}\lambda$
\end{algor}
\item [{else}]~

\begin{algor}[1]
\item [{{*}}] $\Omega\leftarrow\{l'\in c\,|\,\sigma(l')\neq\mbox{false}\}\setminus\{w\}$
\item [{{*}}] /* $\Omega$ is the set of literals that could replace $\neg l$
*/
\item [{if}] $\Omega=\emptyset$ /* no other literal in $c$ can be watched
*/

\begin{algor}[1]
\item [{if}] $\sigma(w)=\mathrm{undef}$ /* $c$ is unit */

\begin{algor}[1]
\item [{{*}}] $\sigma(w)\leftarrow\mbox{true}$/* $w$ is propagated by
$c$ */
\item [{for}] $l\in\mbox{levels}(c)\setminus\{\lambda\}$

\begin{algor}[1]
\item [{{*}}] \textbf{set} $l<_{\Delta}\lambda$
\end{algor}
\item [{endfor}]~
\item [{{*}}] $\Pi\leftarrow\Pi\cup\{w\}$
\end{algor}
\item [{else}]~

\begin{algor}[1]
\item [{{*}}] \textbf{return} $c$ /* $c$ is a conflict */
\end{algor}
\item [{endif}]~
\end{algor}
\item [{else}]~

\begin{algor}[1]
\item [{{*}}] choose $w'\text{\ensuremath{\in}}\Omega$
\item [{{*}}] $\omega(c)\leftarrow\{w,w'\}$ /* $w'$ is watched instead
of $\neg l$ */
\end{algor}
\item [{endif}]~
\end{algor}
\item [{endif}]~
\end{algor}
\item [{endfor}]~
\item [{{*}}] $\Pi\leftarrow\Pi\setminus\{l\}$
\end{algor}
\item [{endwhile}]~
\item [{{*}}] \textbf{return} $\mathrm{NIL}$ /* no conflict occured */
\end{algor}
\caption{\textsc{Propagate}}
\label{Propagate}
\end{algorithm}
Unit clauses are efficiently detected using watched literals~\cite{MWMCFMYZLZSM01},
a method keeping track of two not instantiated literals in each clause
that isn't already satisfied. When a literal $l$ is instantiated,
a clause $c$ cannot become unit unless it contains the opposite literal
$\neg l$ and this literal is watched in $c$. The algorithm thus
only has to check clauses containing $\neg l$ as a watched literal
for possible unit propagations.

A conflict occurs when all literals in a clause are false. CDCL then
infers a conflict clause $\gamma$, which is a logical consequence
of the original formula, is also false under the current assignment
and has only one literal instantiated at the current decision level.
The second largest decision level represented in $\gamma$ is called
the assertion level. The conflict is resolved by undoing all decision
levels above the assertion level. $\gamma$ becomes unit, it is propagated
and the search continues at the assertion level. The algorithm terminates
either when all variables are assigned without causing any conflict
(the formula is satisfied by this assignment) or when a conflict occurs
at decision level 0 (the formula is unsatisfiable).

The pseudocode of PO-CDCL is given in Alg.~\ref{PO-CDCL}. It consists
in a few modifications of the regular CDCL algorithm. A partial order
$\Delta$ keeps track of dependencies between decision levels and
is used to determine the assertion level and the levels to delete
during conflicts. Dependencies are added during the unit propagation
phase detailed in Alg.~\ref{Propagate}. A level $i$ depends on
a level $j$ (noted $j<_{\Delta}i$) when level $j$ had an influence
on unit propagations at level $i$. This can happen in two cases.

First, when a variable $l$ is propagated by a unit clause $c$, this
propagation obviously depends of all other literals in $c$. For all
literals $l'\in c\setminus\{l\}$ whose decision level is different
from the current decision level $\lambda$, the dependency $level(l')<_{\Delta}\lambda$
is added to $\Delta$. This case is handled by lines 14 and 15 of
Alg.~\ref{Propagate}.

Secondly, when a false watched literal $\neg l$ at level $\lambda$
doesn't need to be replaced in a clause $c$ because $w$, the second
watched literal in $c$, is true, the dependency $level(w)<_{\Delta}\lambda$
is added (line 7 of Alg.~\ref{Propagate}). Intuitively, this dependency
means that the true watched literal $w$ avoided a watched literal
replacement at level $\lambda$ and therefore had an impact on the
unit propagations at this level. More technically, this dependency
ensures that the clause will remain correctly watched by forbidding
to uninstantiate $w$ while keeping $\neg l$ instantiated.

With a partial order on decision levels, the backtrack phase only
requires to delete levels that depend on the assertion level (and
of course the conflict level itself). This deletion (at lines 12 and
13 of Alg.~\ref{PO-CDCL}) is necessary to keep the consistency of
the algorithm by preventing circular dependencies between levels.

Finally, the last modification affects the definition of the assertion
level. This level has to be involved in the conflict clause, and the
conflict clause must become unit after the backtrack. This implies
that no decision level occuring in the conflict clause must be undone
by the backtrack, except for the conflict level $\lambda$. In a total
order CDCL, the assertion level is uniquely defined as the second
largest decision level in the conflict clause. With a partial order,
however, any decision level in the conflict clause can be chosen as
the assertion level, provided that no other level involved in the
conflict, except $\lambda$, depends on it. In other words, the assertion
level can be any maximal element of $<_{\Delta}$ restricted to the
set of conflict clause levels different from $\lambda$ (lines 10
and 11 of Alg.~\ref{PO-CDCL}).

Similarly to the original CDCL algorithm, PO-CDCL is complete, correct
and always terminates~\cite{AMRV12}.

\section{Related Works\label{sec:Related-Works}}

PO-CDCL is conceptually related to some variations of the Conflict-Direct
Backjumping (CBJ) algorithm for CSP solving which, similarly to CDCL
for SAT, resolves conflicts by computing a nogood (equivalent of the
conflict clause) and deleting the entire search progress starting
at the culprit variable decision (roughly equivalent of the decision
at the conflict level). In the case of CSPs, search progress consists
not only of variable assignments, but also of values eliminated from
domains of variables.

Dynamic Backtracking (DB)~\cite{MLG93}, in contrast with CBJ, only
undoes the culprit variable and restores only eliminated values for
which the culprit variable was part of the nogood. This strategy is
equivalent to dynamically moving the culprit variable to the end of
the search branch before undoing it, provided a limited amount of
search information is deleted. It has the advantage of only partially
undoing the work made after the culprit variable. Similarly to PO-CDCL,
it minimizes the quantity of undone search progress by relaxing the
strict total order on variable decisions. The main difference is that
DB is defined as a search-only algorithm without any inference; therefore
the conflict can always be resolved without undoing any other decision
than the culprit variable. Also note that the usual total order is
considered during the analysis phase; unlike the assertion level in
PO-CDCL, the culprit variable in DB remains thus uniquely defined.

Partial Order Backtracking (POB)~\cite{DAMA93} similarly only uninstantiates
the culprit variable for each conflict and only restores values whose
elimination depended on it. The difference is that it initially allows
to choose the culprit variable amongst all variables in the nogood,
but progressively sets precedence constraints between variables in
order to ensure termination. This freedom in the choice of the culprit
variable is stronger than the freedom PO-CDCL offers for choosing
the assertion level. It however however comes with a strong permanent
and increasing constraint on decision heuristics, whereas constraints
set by PO-CDCL between decision levels only apply until these decision
levels are undone, and hence have no impact on the choice of decision
variables.

Tree decompositions methods integrated within CDCL~\cite{JHAD03,WLPvB04,VDPK04,AMRV10}
and CBJ~\cite{PJCT03} solvers also indirectly limit the quantity
of unrelated instantiations undone during a backtrack. Decompositions~\cite{NRPDS86}
are used to compute recursive separators of the instance, i.e. sets
of variables whose instantiation breaks the problem in several connected
components. These methods start the search by instantiating all separator
variables, and then completely instantiate a connected component before
making any decision in another component. When a conflict occurs in
a connected component, the resulting backtrack then can't destroy
any part of the search in other components thanks to this constrained
ordering. Besides scalability issues which make it very difficult
to efficiently compute useful decompositions on large SAT problems~\cite{AMRV10},
tree decompositions only capture the static connectivity of a problem
and therefore can't take into account the polarity of instantiations
and the many propagations they cause. At any point of the search,
the actual connectivity is likely to be much more sparse than predicted
by decompositions. Therefore, a conflict in a connected component
may actually delete instantiations in another component. PO-CDCL,
on the other hand, considers the exact connectivity at any time of
the search. It also distinguishes sets of variables that haven't interacted
yet in the current search branch even if they belong to the same connected
component; it considers actual interactions between already instantiated
variables rather than potential interactions between still unassigned
variables.

Finally, phase saving~\cite{KPAD07}, in contrast with tree decompositions,
is a very lightweight approach. It simply memorizes the last polarity
assigned to a variable and reuses it if the variable is picked for
a decision. Phase saving actually doesn't prevent instantiations from
being undone, but makes it possible to rediscover the deleted instantiations
later. It thus allows to recover search progress that was lost during
a conflict resolution. However, unlike partial order CDCL, this recovery
doesn't save the computational cost of repeating the time-consuming
propagation phase. Also, phase saving memorizes the polarity of all
variables, even if they were actually involved in the conflict. This
side effect sometimes decreases solving performance, as reported by
the authors~\cite{KPAD07}.

Note that, at the opposite, some strategies have been designed to
enhance SAT solving by increasing the quantity of instantiations undone
during conflict-directed backtracks~\cite{ANVR10,ABILJTdSJMS05}.

\section{PO-CDCL Analysis and Assertion Level Heuristics\label{sec:Assertion-level-heuristics}}

The PO-CDCL algorithm was implemented by introducing a partial order
on decision levels in the state-of-the-art CDCL solver \textsc{Glucose}
1.0~\cite{GALS09}. The resulting PO-CDCL solver is named \textsc{PO-Glucose}
and its source code is available at \urldef{\POCDCL}\url{http://www.info2.uqam.ca/~villemaire_r/Recherche/SAT/120619 generalized_glucose.tar.gz}\POCDCL.
In this implementation, level dependencies are stored in three structures:
two directed adjacency lists, representing the relation in both directions,
and one boolean matrix. The combination of these structures allows
to perform efficiently all operations on the partial relation: somes
cases require to check the relation between a precise pair of decision
levels, which can be done in constant time using the matrix. At the
opposite, the algorithm sometimes requires to list of all levels depending
on a given level, in which case using the adjacency list is obviously
more efficiently, particularly when there are many active decision
levels with little interdependence. Note that only direct dependencies
are stored; transitive dependencies are only needed during conflict
resolution on a small subset of variables and it is much more efficient
to compute this partial transitive closure when it is required than
to enforce and store transitivity during the entire algorithm.

As the size of the matrix grows quadratically with the number of decision
levels, our implementation disables it if this number reaches a predefinite
threshold. The algorithm then proceeds using only adjacency lists,
which is slightly less efficient but significantly better than exhausting
primary memory. Theoretically, the size of adjacency lists could also
grow quadratically in the case of dense dependencies between decision
levels; however, it seems that in practice the number of decision
levels tends to decrease when this density grows. The memory requirement
of adjacency lists thus remains relatively moderate.

The remaining of this section presents and compares experimental results
obtained with this implementation and with the original \textsc{Glucose}
solver. We will more particularly focus on the impact of assertion
level choice heuristics on the overall behaviour and performance of
the algorithm. All tests were run on a 3.16 GHz Intel Core 2 Duo CPU
with 3 GB of RAM, running a Ubuntu 11.10 OS, with a time limit of
1 hour for each execution (not including the preprocessing phase,
which is identical for all tested variants).

We previously noticed~\cite{AMRV12} that since PO-CDCL is designed
to take advantage of the independence between decision levels during
solving, it performs best on problems where this independence is relatively
high. If we consider the partial order $\Delta$ as a set of ordered
pairs of decision levels, such that the first level in each pair depends
of the second level of the pair, the cardinality of $\Delta$ can
be used as a measure of this independence. Benchmarks from formal
verification of superscalar microprocessors~\cite{MNVREB03} are
an example of problems with a very sparse relationship between levels,
possibly because of the high parallelism in verified models. Therefore,
the following experiments were conducted on 6 series of these benchmarks:
\begin{itemize}
\item \textsf{pipe\_unsat\_1.0} and \textsf{pipe\_unsat\_1.1} verify correct
specifications of various-sized superscalar microprocessors with two
different encoding variants;
\item \textsf{pipe\_sat\_1.0} and \textsf{pipe\_sat\_1.1} represent ten
different buggy variants of the size 12 case, again encoded in two
different ways;
\item \textsf{pipe\_ooo\_unsat\_1.0} and \textsf{pipe\_ooo\_unsat\_1.1}
are two different encodings verifying the correctness of various-sized
superscalar microprocessors handling out-of-order execution of instructions.
\end{itemize}
\begin{sidewaystable}
\begin{centering}
{\scriptsize }%
\begin{tabular}{|c|c|r|r|r|r|r|r|r|r|r|r|r|r|r|r|}
\hline 
\multirow{2}{*}{{\scriptsize family}} & \multirow{2}{*}{{\scriptsize \#inst}} & \multicolumn{2}{c|}{{\scriptsize TO}} & \multicolumn{2}{c|}{{\scriptsize TO-phase}} & \multicolumn{2}{c|}{{\scriptsize PO}} & \multicolumn{2}{r|}{{\scriptsize PO-least-undos}} & \multicolumn{2}{r|}{{\scriptsize PO-most-undos}} & \multicolumn{2}{r|}{{\scriptsize PO-least-deps}} & \multicolumn{2}{r|}{{\scriptsize PO-most-deps}}\tabularnewline
\cline{3-16} 
 &  & {\scriptsize \#to} & {\scriptsize time} & {\scriptsize \#to} & {\scriptsize time} & {\scriptsize \#to} & {\scriptsize time} & {\scriptsize \#to} & {\scriptsize time} & {\scriptsize \#to} & {\scriptsize time} & {\scriptsize \#to} & {\scriptsize time} & {\scriptsize \#to} & {\scriptsize time}\tabularnewline
\hline 
\textsf{\scriptsize pipe\_sat\_1.0} & {\scriptsize 10} & {\scriptsize 6} & {\scriptsize 25~364} & {\scriptsize 0} & {\scriptsize 6~887} & {\scriptsize 0} & {\scriptsize 6~601} & {\scriptsize 0} & {\scriptsize 7~334} & {\scriptsize 0} & {\scriptsize 2~042} & {\scriptsize 0} & {\scriptsize 1~264} & {\scriptsize 2} & {\scriptsize 10~399}\tabularnewline
\hline 
\textsf{\scriptsize pipe\_sat\_1.1} & {\scriptsize 10} & {\scriptsize 1} & {\scriptsize 7~258} & {\scriptsize 0} & {\scriptsize 1~182} & {\scriptsize 1} & {\scriptsize 3~766} & {\scriptsize 1} & {\scriptsize 4~010} & {\scriptsize 0} & {\scriptsize 186} & {\scriptsize 0} & {\scriptsize 185} & {\scriptsize 1} & {\scriptsize 3~820}\tabularnewline
\hline 
\textsf{\scriptsize pipe\_unsat\_1.0} & {\scriptsize 13} & {\scriptsize 5} & {\scriptsize 23~172} & {\scriptsize 7} & {\scriptsize 25~627} & {\scriptsize 5} & {\scriptsize 19~456} & {\scriptsize 5} & {\scriptsize 19~697} & {\scriptsize 5} & {\scriptsize 19~192} & {\scriptsize 5} & {\scriptsize 20~338} & {\scriptsize 4} & {\scriptsize 17~742}\tabularnewline
\hline 
\textsf{\scriptsize pipe\_unsat\_1.1} & {\scriptsize 14} & {\scriptsize 5} & {\scriptsize 20~706} & {\scriptsize 7} & {\scriptsize 28~460} & {\scriptsize 6} & {\scriptsize 23~591} & {\scriptsize 6} & {\scriptsize 23~130} & {\scriptsize 6} & {\scriptsize 22~837} & {\scriptsize 6} & {\scriptsize 23~149} & {\scriptsize 6} & {\scriptsize 22~198}\tabularnewline
\hline 
\textsf{\scriptsize pipe\_ooo\_unsat\_1.0} & {\scriptsize 9} & {\scriptsize 2} & {\scriptsize 10~757} & {\scriptsize 1} & {\scriptsize 6~989} & {\scriptsize 2} & {\scriptsize 11~670} & {\scriptsize 3} & {\scriptsize 13~321} & {\scriptsize 2} & {\scriptsize 10~613} & {\scriptsize 2} & {\scriptsize 10~799} & {\scriptsize 2} & {\scriptsize 10~420}\tabularnewline
\hline 
\textsf{\scriptsize pipe\_ooo\_unsat\_1.1} & {\scriptsize 10} & {\scriptsize 1} & {\scriptsize 11~457} & {\scriptsize 4} & {\scriptsize 30~563} & {\scriptsize 1} & {\scriptsize 12~153} & {\scriptsize 2} & {\scriptsize 16~185} & {\scriptsize 2} & {\scriptsize 15~909} & {\scriptsize 2} & {\scriptsize 16~485} & {\scriptsize 1} & {\scriptsize 11~592}\tabularnewline
\hline 
{\scriptsize total} & {\scriptsize 66} & {\scriptsize 20} & {\scriptsize 98~714} & {\scriptsize 19} & {\scriptsize 99~708} & {\scriptsize 15} & {\scriptsize 77~237} & {\scriptsize 17} & {\scriptsize 83~677} & {\scriptsize 15} & {\scriptsize 70~870} & {\scriptsize 15} & {\scriptsize 72~236} & {\scriptsize 16} & {\scriptsize 76~196}\tabularnewline
\hline 
\end{tabular}\caption{Compared performances of \textsc{Glucose} without (\emph{TO}) and
with (\emph{TO-phase}) phase saving, PO-\textsc{Glucose} with the
default chronological assertion level heuristic (\emph{PO}) and with
4 other heuristics based on the amount of instantiations undone by
the backtrack (\emph{PO-least-undos}, \emph{PO-most-undos}) or on
the number of level dependencies added (\emph{PO-least-deps}, \emph{PO-most-deps}).
For each \emph{series} of benchmarks, containing \emph{\#inst} instances,
the number of timeouts (\emph{\#to}) and the total solving time in
seconds (\emph{time}) is given.}
{\scriptsize \label{table1}\bigskip{}
}
\par\end{centering}{\scriptsize \par}

\begin{centering}
{\scriptsize }%
\begin{tabular}{|c|c|r|r|r|r|r|r|r|r|r|r|r|r|r|r|}
\hline 
\multirow{2}{*}{{\scriptsize series}} & \multirow{2}{*}{{\scriptsize \#inst}} & \multicolumn{2}{c|}{{\scriptsize TO}} & \multicolumn{2}{c|}{{\scriptsize TO-phase}} & \multicolumn{2}{c|}{{\scriptsize PO}} & \multicolumn{2}{r|}{{\scriptsize PO-least-undos}} & \multicolumn{2}{r|}{{\scriptsize PO-most-undos}} & \multicolumn{2}{r|}{{\scriptsize PO-least-deps}} & \multicolumn{2}{r|}{{\scriptsize PO-most-deps}}\tabularnewline
\cline{3-16} 
 &  & {\scriptsize \#to} & {\scriptsize checks} & {\scriptsize \#to} & {\scriptsize checks} & {\scriptsize \#to} & {\scriptsize checks} & {\scriptsize \#to} & {\scriptsize checks} & {\scriptsize \#to} & {\scriptsize checks} & {\scriptsize \#to} & {\scriptsize checks} & {\scriptsize \#to} & {\scriptsize checks}\tabularnewline
\hline 
\textsf{\scriptsize pipe\_sat\_1.0} & {\scriptsize 10} & {\scriptsize 6} & {\scriptsize 60~962} & {\scriptsize 0} & {\scriptsize 174~962} & {\scriptsize 0} & {\scriptsize 74~034} & {\scriptsize 0} & {\scriptsize 104~876} & {\scriptsize 0} & {\scriptsize 23~970} & {\scriptsize 0} & {\scriptsize 12~816} & {\scriptsize 2} & {\scriptsize 45~065}\tabularnewline
\hline 
\textsf{\scriptsize pipe\_sat\_1.1} & {\scriptsize 10} & {\scriptsize 1} & {\scriptsize 257~229} & {\scriptsize 0} & {\scriptsize 38~492} & {\scriptsize 1} & {\scriptsize 1~438} & {\scriptsize 1} & {\scriptsize 4~542} & {\scriptsize 0} & {\scriptsize 1~361} & {\scriptsize 0} & {\scriptsize 1~123} & {\scriptsize 1} & {\scriptsize 1~938}\tabularnewline
\hline 
\textsf{\scriptsize pipe\_unsat\_1.0} & {\scriptsize 13} & {\scriptsize 5} & {\scriptsize 204~171} & {\scriptsize 7} & {\scriptsize 336~799} & {\scriptsize 5} & {\scriptsize 34~804} & {\scriptsize 5} & {\scriptsize 39~208} & {\scriptsize 5} & {\scriptsize 23~256} & {\scriptsize 5} & {\scriptsize 52~394} & {\scriptsize 4} & {\scriptsize 58~161}\tabularnewline
\hline 
\textsf{\scriptsize pipe\_unsat\_1.1} & {\scriptsize 14} & {\scriptsize 5} & {\scriptsize 95~137} & {\scriptsize 7} & {\scriptsize 384~249} & {\scriptsize 6} & {\scriptsize 51~028} & {\scriptsize 6} & {\scriptsize 35~829} & {\scriptsize 6} & {\scriptsize 26~370} & {\scriptsize 6} & {\scriptsize 34~717} & {\scriptsize 6} & {\scriptsize 11~874}\tabularnewline
\hline 
\textsf{\scriptsize pipe\_ooo\_unsat\_1.0} & {\scriptsize 9} & {\scriptsize 2} & {\scriptsize 124~488} & {\scriptsize 1} & {\scriptsize 107~063} & {\scriptsize 2} & {\scriptsize 75~783} & {\scriptsize 3} & {\scriptsize 48~229} & {\scriptsize 2} & {\scriptsize 55~363} & {\scriptsize 2} & {\scriptsize 56~183} & {\scriptsize 2} & {\scriptsize 51~501}\tabularnewline
\hline 
\textsf{\scriptsize pipe\_ooo\_unsat\_1.1} & {\scriptsize 10} & {\scriptsize 1} & {\scriptsize 141~491} & {\scriptsize 4} & {\scriptsize 57~129} & {\scriptsize 1} & {\scriptsize 76~962} & {\scriptsize 2} & {\scriptsize 31~329} & {\scriptsize 2} & {\scriptsize 25~691} & {\scriptsize 2} & {\scriptsize 35~458} & {\scriptsize 1} & {\scriptsize 66~023}\tabularnewline
\hline 
{\scriptsize total} & {\scriptsize 66} & {\scriptsize 20} & {\scriptsize 740~730} & {\scriptsize 19} & {\scriptsize 1~098~693} & {\scriptsize 15} & {\scriptsize 314~050} & {\scriptsize 17} & {\scriptsize 264~013} & {\scriptsize 15} & {\scriptsize 156~011} & {\scriptsize 15} & {\scriptsize 192~692} & {\scriptsize 16} & {\scriptsize 234~562}\tabularnewline
\hline 
\end{tabular}
\par\end{centering}{\scriptsize \par}

\caption{Compared performances of the same \textsc{Glucose} and PO-\textsc{Glucose}
variations on the same series of benchmarks. For each series, besides
the number of timeouts (\emph{\#to}), the total number of clause checks
performed (\emph{checks}, given in millions) is listed. When several
solvers timed out on the same instance, they were considered as having
all needed the same amount of clause checks (the smallest amount amongst
timed out solvers).}
{\scriptsize \label{table 2}}
\end{sidewaystable}
Benchmarks verifying correct and buggy specifications are respectively
unsatisfiable and satisfiable.

\textsc{Glucose} implements the phase saving strategy mentioned in
section~\ref{sec:Related-Works}. We disabled phase saving in PO-\textsc{Glucose}
because partial order CDCL was partly designed as an alternative to
phase saving. Moreover, preliminary tests indicated that PO-\textsc{Glucose}
often performs significantly better with phase saving disabled. To
make sure the performance differences we observe are not simply caused
by the presence or abscence of phase saving rather than by the partial
order, we compared \textsc{PO-Glucose} with the original \textsc{Glucose},
but also with a variant in which phase saving is disabled.

In \textsc{PO-Glucose}, the partial order management causes a significant
calculation overhead during solving. Indeed, each propagation requires
to check and possibly add several level dependencies. As a result,
given the same execution time on the same instance, PO-Glucose performs
on average about 40\% less clause checks (i.e. the number of executions
of the innermost\textbf{ for} loop at lines 4 to 21 of Alg.~\ref{Propagate})
than \textsc{Glucose}. We think this overhead can't be significantly
reduced unless we find some lazy strategy to manage dependencies.
Therefore, besides the CPU time used to solve each instance, we also
report the number of clauses checked for possible propagations during
solving. This quantity gives some insight about which proportion of
the PO-\textsc{Glucose} solving time is spent in the search itself
and to what extent this time is due to dependency management.

\subsection{Analyzing efficiency of PO-CDCL}

In this subsection, we will consider the default version of PO-\textsc{Glucose}
as described in~\cite{AMRV12} with a choice of the assertion level
similar to its definition in a total order CDCL: amongst all candidate
assertion levels, the most recently created one is picked. This default
version is named \emph{PO} in all tables and figures of this paper.
Results of \textsc{Glucose} with and without phase saving are labelled
as \emph{TO-phase} and \emph{TO} respectively, \emph{TO-phase} being
the default \textsc{Glucose} setting.

As expected, when a conflict occurs during a CDCL solving, there is
in practice often a non-negligible quantity of instantiations between
the assertion level and the conflict level. Therefore, on our formal
verification instances, PO-\textsc{Glucose} locally saves on average
15\% of instantiations that would be deleted by a regular CDCL algorithm
(they are located in decision levels instantiated after the assertion
level but not depending on it). If we consider an entire solving trace,
it however deletes on average approximately the same number of instantiations
per conflict than the original \textsc{Glucose}, as shown in Table~\ref{table3}.
The efficiency of PO-\textsc{Glucose} is thus not obtained by accumulating
instantiations faster than with a total order; saved instantiations
are likely to be deleted later. However, we will show that although
instantiations are only saved temporarily, they can have a significant
impact on the overall search.

\begin{sidewaysfigure}
\begin{tabular}{cc}
\includegraphics[scale=0.7]{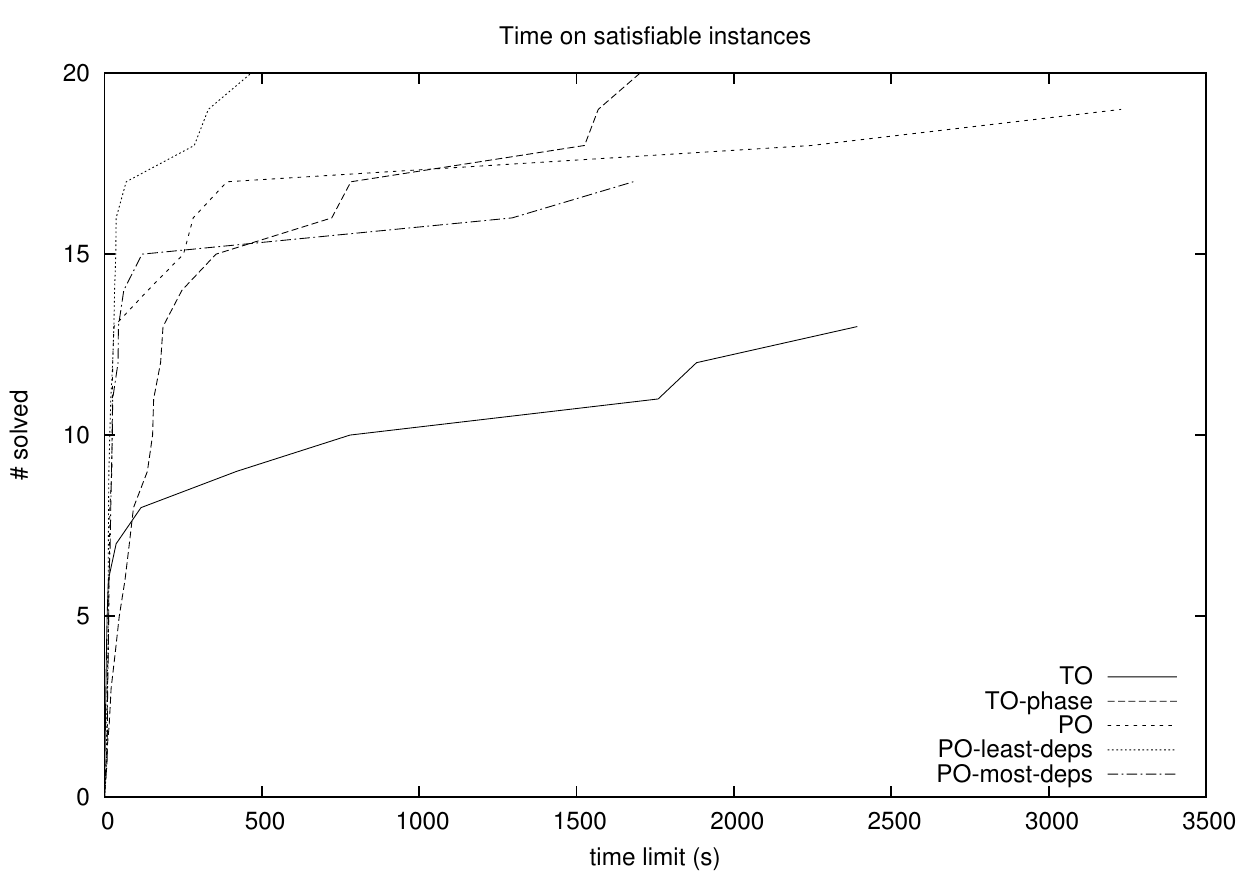} & \includegraphics[scale=0.7]{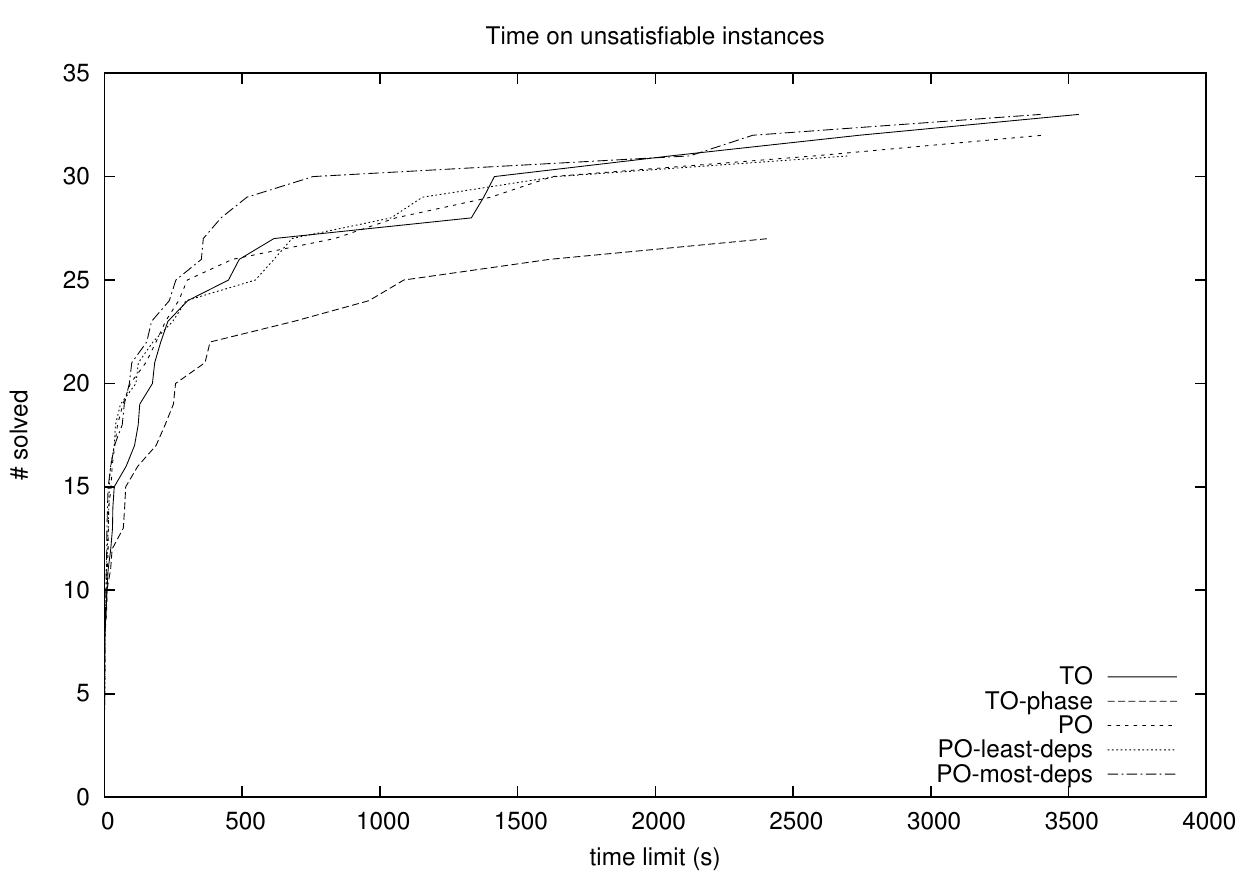}\tabularnewline
\includegraphics[scale=0.7]{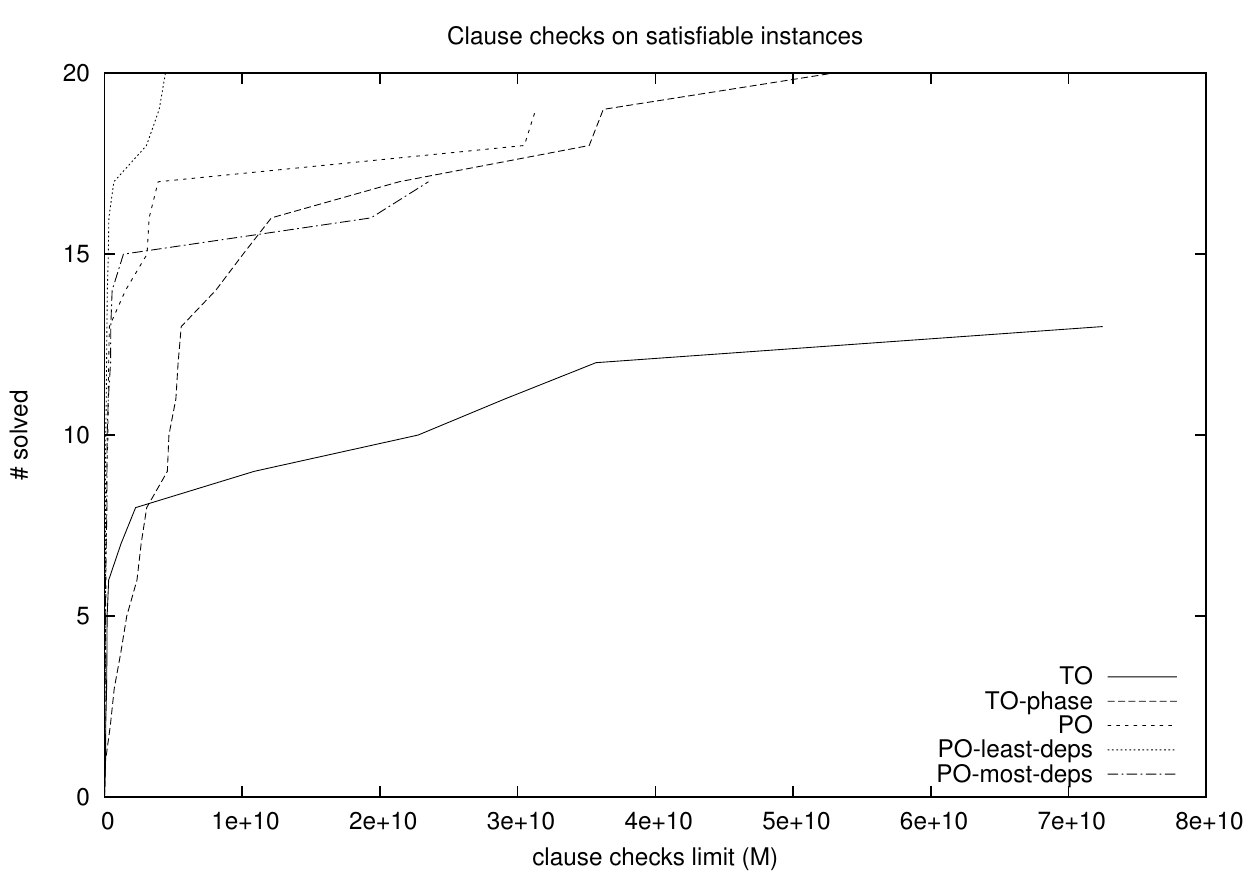} & \includegraphics[scale=0.7]{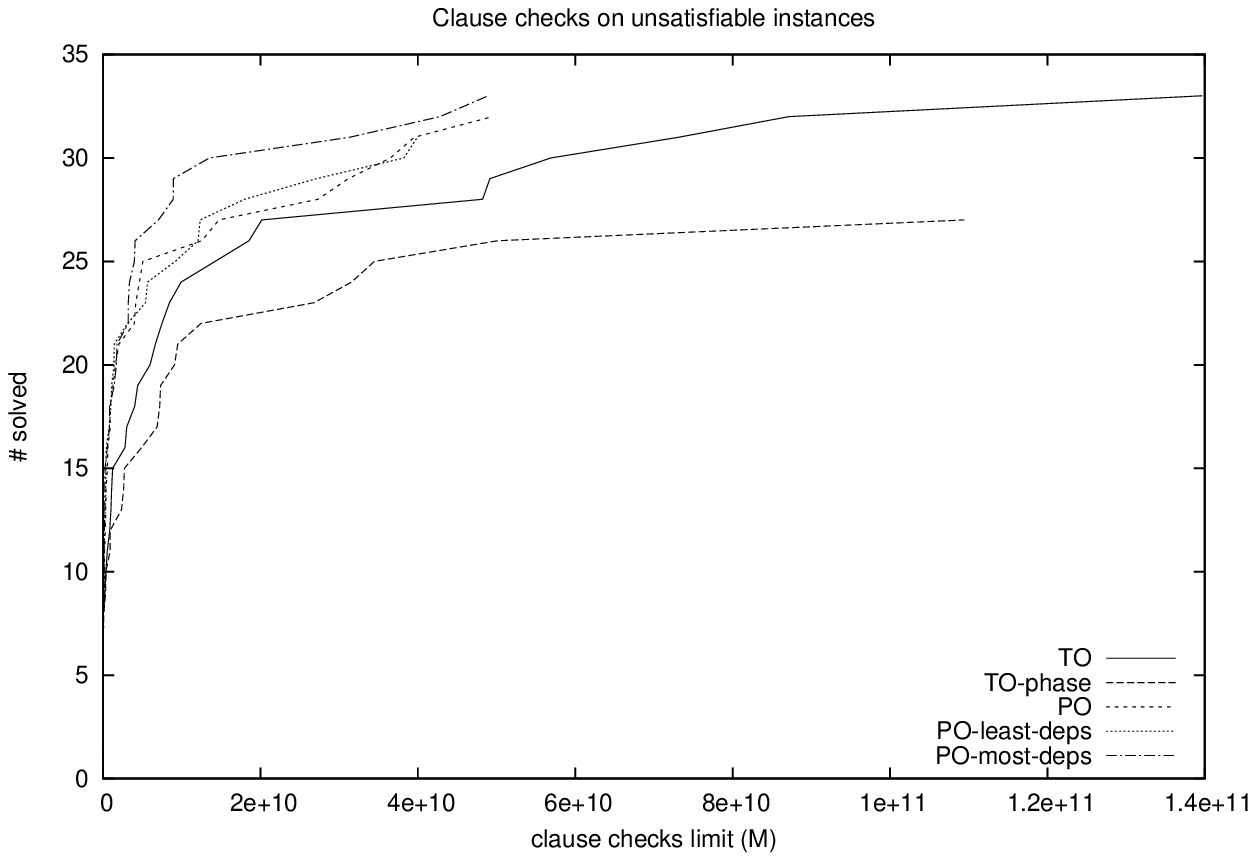}\tabularnewline
\end{tabular}

\caption{Four cactus plots comparing the performances of two variants of \textsc{Glucose},
without (\emph{TO}) and with \emph{(TO-phase}) phase saving, with
the default PO-CDCL (\emph{PO}) and PO-CDCL with two dependency-counting
heuristics (\emph{PO-least-deps, PO-most-deps}). Left plots compare
the algorithms on the 20 satisfiable instances, right plots on the
46 unsatisfiable instances. x-axis measures the solving time on top
plots and the clause checks performed on bottom plots.}
\label{cactus-plot}
\end{sidewaysfigure}
Tables~\ref{table1} and \ref{table 2} show respectively the total
time and clause checks needed to solve each benchmark family with
this chronological heuristic, compared with performances of the two
total order variants. Both versions of \textsc{Glucose} have very
contrasted results: the default version with phase saving clearly
outperforms the version without phase saving on both satisfiable series,
but conversely the version without phase saving performs better on
3 of the 4 unsatisfiable families.

When comparing solving time for each series separately, the performance
of PO-\textsc{Glucose} is generally close to the best performing \textsc{Glucose}
version and significantly better than the other (except on \textsf{pipe\_ooo\_unsat\_1.0},
where it requires a little more time than the slowest \textsc{Glucose}
variant). It also never causes more than one additional timeout than
the best performing \textsc{Glucose} version. Thanks to this more
balanced behaviour, it significantly outperforms both \textsc{Glucose}
with and without phase saving when considering the total solving time
on all benchmarks, and manages to solve 4 to 5 more instances in the
given time limit.

The cactus plots of Fig.~\ref{cactus-plot} give a better view of
the performances on individual instances. Top figures show how many
satisfiable and unsatisfiable instances respectively can be solved
within a given time limit. The top left figure indicates that PO-\textsc{Glucose}
manages to solve many instances very quickly (13 out of 20 are solved
in less than 30 seconds each). When the time limit increases, it is
eventually beaten by the default setup of \textsc{Glucose} which is
able to solve the 3 most difficult instances in a little less than
30 minutes while PO-\textsc{Glucose} needs more time and fails to
solve one of them within one hour. It however easily outperforms \textsc{Glucose}
without phase saving. 

On unsatisfiable instances (top right figure), PO-Glucose considerably
outperforms the default version of \textsc{Glucose} with phase saving
enabled, no matter what time limit is considered. Within one hour,
it solves 5 more instances than default \textsc{Glucose}. \textsc{Glucose}
without phase saving is however more efficient and slightly outperforms
PO-\textsc{Glucose} on high time limits, although the latter manages
to solve more instances in 500 seconds or less.

Since about 40\% of the solving time is an overhead due to handling
level dependencies, the performance of PO-\textsc{Glucose} is even
better when the number of clause checks is considered. Bottom plots
of Fig.~\ref{cactus-plot} indicate that PO-\textsc{Glucose} significantly
outperforms both versions of \textsc{Glucose} on most satisfiable
and unsatisfiable instances. This observation is confirmed by Table~\ref{table 2},
which shows that PO-\textsc{Glucose} often requires dramatically less
clause checks to solve the same amount of instances than \textsc{Glucose}.
Overall, PO-\textsc{Glucose} solves more instances than both \textsc{Glucose}
implementations with twice to thrice less clause checks.

This efficiency can be explained by the effect of saved instantiations
on the search. Table~\ref{table4} shows the average amount of clause
checks necessary to reach a conflict for various \textsc{Glucose}
and PO-\textsc{Glucose} versions. This quantity is almost always dramatically
lowered by PO-\textsc{Glucose}, no matter what assertion heuristic
is used. The instantiations saved by partial order backtracks, even
if they are eventually deleted, seem to be often relevant and help
reaching conflicts much faster. As each conflict prunes a part of
the search space, partial order thus apparently dramatically improves
this pruning, which obviously should help in proving unsatisfiability
faster, but in also guiding the search in satisfiable instances towards
branches of the search space containing models.

\subsection{Assertion level heuristics}

The default chronological assertion level choice used in the previous
subsection was designed to remain as close as possible to the original
CDCL algorithm and evaluate the efficiency gain that can be obtained
solely by removing less instantiations during backtracks, without
further modifying the search. However, we will show that this choice
can significantly modify the way the search space is explored, and
that particular heuristics can be used to further improve performances
of \textsc{PO-Glucose}.

\begin{sidewaystable}
\begin{centering}
{\footnotesize }%
\begin{tabular}{|c|c|r|r|r|r|r|r|r|}
\hline 
\multirow{1}{*}{{\footnotesize family}} & \multirow{1}{*}{{\footnotesize \#inst}} & \multicolumn{1}{c|}{{\footnotesize TO}} & \multicolumn{1}{c|}{{\footnotesize TO-phase}} & \multicolumn{1}{c|}{{\footnotesize PO}} & \multicolumn{1}{c|}{{\footnotesize PO-least-undos}} & \multicolumn{1}{c|}{{\footnotesize PO-most-undos}} & \multicolumn{1}{c|}{{\footnotesize PO-least-deps}} & \multicolumn{1}{c|}{{\footnotesize PO-most-deps}}\tabularnewline
\hline 
\textsf{\footnotesize pipe\_sat\_1.0} & {\footnotesize 10} & {\footnotesize 1~226} & {\footnotesize 2~053} & {\footnotesize 1~751} & {\footnotesize 1~698} & {\footnotesize 3~680} & {\footnotesize 4~602} & {\footnotesize 1~972}\tabularnewline
\hline 
\textsf{\footnotesize pipe\_sat\_1.1} & {\footnotesize 10} & {\footnotesize 1~099} & {\footnotesize 1~726} & {\footnotesize 1~957} & {\footnotesize 1~599} & {\footnotesize 3~834} & {\footnotesize 4~983} & {\footnotesize 1~691}\tabularnewline
\hline 
\textsf{\footnotesize pipe\_unsat\_1.0} & {\footnotesize 13} & {\footnotesize 885} & {\footnotesize 968} & {\footnotesize 1~050} & {\footnotesize 1~046} & {\footnotesize 1~241} & {\footnotesize 1~244} & {\footnotesize 970}\tabularnewline
\hline 
\textsf{\footnotesize pipe\_unsat\_1.1} & {\footnotesize 14} & {\footnotesize 1~124} & {\footnotesize 1~054} & {\footnotesize 1~096} & {\footnotesize 1~066} & {\footnotesize 1~284} & {\footnotesize 1~316} & {\footnotesize 974}\tabularnewline
\hline 
\textsf{\footnotesize pipe\_ooo\_unsat\_1.0} & {\footnotesize 9} & {\footnotesize 648} & {\footnotesize 679} & {\footnotesize 660} & {\footnotesize 648} & {\footnotesize 685} & {\footnotesize 693} & {\footnotesize 624}\tabularnewline
\hline 
\textsf{\footnotesize pipe\_ooo\_unsat\_1.1} & {\footnotesize 10} & {\footnotesize 784} & {\footnotesize 610} & {\footnotesize 749} & {\footnotesize 718} & {\footnotesize 781} & {\footnotesize 755} & {\footnotesize 741}\tabularnewline
\hline 
{\footnotesize average} & {\footnotesize 11} & {\footnotesize 972} & {\footnotesize 1~172} & {\footnotesize 1~204} & {\footnotesize 1~129} & {\footnotesize 1~867} & {\footnotesize 2~185} & {\footnotesize 1~151}\tabularnewline
\hline 
\end{tabular}
\par\end{centering}{\footnotesize \par}

\caption{Comparison of the average number of instantiations undone at each
backtrack by various solvers on some benchmark series. Solvers and
benchmarks tested are the same as in Table~\ref{table1}. }
\label{table3}{\footnotesize \bigskip{}
}{\footnotesize \par}

\centering{}{\footnotesize }%
\begin{tabular}{|c|c|r|r|r|r|r|r|r|}
\hline 
\multirow{1}{*}{{\footnotesize series}} & \multirow{1}{*}{{\footnotesize \#inst}} & \multicolumn{1}{c|}{{\footnotesize TO}} & \multicolumn{1}{c|}{{\footnotesize TO-phase}} & \multicolumn{1}{c|}{{\footnotesize PO}} & \multicolumn{1}{c|}{{\footnotesize PO-least-undos}} & \multicolumn{1}{c|}{{\footnotesize PO-most-undos}} & \multicolumn{1}{c|}{{\footnotesize PO-least-deps}} & \multicolumn{1}{c|}{{\footnotesize PO-most-deps}}\tabularnewline
\hline 
\textsf{\footnotesize pipe\_sat\_1.0} & {\footnotesize 10} & {\footnotesize 5~164} & {\footnotesize 276} & {\footnotesize 18} & {\footnotesize 23} & {\footnotesize 13} & {\footnotesize 9} & {\footnotesize 37}\tabularnewline
\hline 
\textsf{\footnotesize pipe\_sat\_1.1} & {\footnotesize 10} & {\footnotesize 2~999} & {\footnotesize 16} & {\footnotesize 11} & {\footnotesize 17} & {\footnotesize 6} & {\footnotesize 10} & {\footnotesize 25}\tabularnewline
\hline 
\textsf{\footnotesize pipe\_unsat\_1.0} & {\footnotesize 13} & {\footnotesize 1~214} & {\footnotesize 1~499} & {\footnotesize 21} & {\footnotesize 28} & {\footnotesize 18} & {\footnotesize 10} & {\footnotesize 39}\tabularnewline
\hline 
\textsf{\footnotesize pipe\_unsat\_1.1} & {\footnotesize 14} & {\footnotesize 203} & {\footnotesize 1~271} & {\footnotesize 19} & {\footnotesize 21} & {\footnotesize 13} & {\footnotesize 9} & {\footnotesize 19}\tabularnewline
\hline 
\textsf{\footnotesize pipe\_ooo\_unsat\_1.0} & {\footnotesize 9} & {\footnotesize 99} & {\footnotesize 13} & {\footnotesize 7} & {\footnotesize 8} & {\footnotesize 5} & {\footnotesize 4} & {\footnotesize 7}\tabularnewline
\hline 
\textsf{\footnotesize pipe\_ooo\_unsat\_1.1} & {\footnotesize 10} & {\footnotesize 25} & {\footnotesize 1~283} & {\footnotesize 9} & {\footnotesize 9} & {\footnotesize 6} & {\footnotesize 6} & {\footnotesize 8}\tabularnewline
\hline 
{\footnotesize average} & {\footnotesize 11} & {\footnotesize 1~546} & {\footnotesize 805} & {\footnotesize 14} & {\footnotesize 19} & {\footnotesize 11} & {\footnotesize 8} & {\footnotesize 23}\tabularnewline
\hline 
\end{tabular}\caption{Comparison of the average number of clause checks (in millions) needed
to reach a conflict by various solvers on some benchmark series. Solvers
and benchmarks tested are the same as in Table~\ref{table1}. Note
that the correlation between solving performances and the number of
clause checks per conflict can be confirmed by comparing both total
order versions \emph{TO} and \emph{TO-phase}: the best performing
version on a benchmark series is always the one with the least checks
per conflict.}
\label{table4}
\end{sidewaystable}
Tests with the chronological assertion level choice showed that in
31\% of the conflicts, there are several candidate assertion levels,
and when it happens there are on average about 10 distinct candidate
levels. The strategy used to choose the assertion level thus can potentially
have a significant impact on the entire search. Since the primary
goal of PO-CDCL is to save instantiations during backtracks, a straightforward
local heuristic (named \emph{PO-less-undos} in tables) consists in
picking the candidate assertion level that will undo the least instantiations,
i.e. that minimizes the quantity of variables located in decision
levels depending on the candidate assertion level. However, according
to Table~\ref{table3}, this strategy almost doesn't modify the average
number of undos per conflict. Consequently, performances obtained
with this heuristic are relatively close to results of the default
chronological heuristic, as shown in Tables~\ref{table1} and~\ref{table 2}.
This seems to indicate that the chronological heuristic already often
picks assertion levels causing few uninstantiations.

Surprisingly, the opposite heuristic of picking the assertion level
that will cause the most deletions (\emph{PO-most-undos}) is much
more interesting. Its performances on unsatisfiable instances are
very close to performances of the chronological heuristic. However,
as shown in Tables~\ref{table1} and~\ref{table 2}, it dramatically
reduces the time and clause checks needed to solve satisfiable instances.
\textsf{pipe\_sat\_1.0} is solved about 3 times faster and with 7
times less clause checks than the best performing \textsc{Glucose}
version. \textsf{pipe\_sat\_1.1} is solved more than 6 times faster
and with 28 times less clause checks.

On these satisfiable series, as indicated by Table~\ref{table3},
the \emph{most undos} heuristic deletes about twice more instantiations
than default\textsc{ PO-Glucose} and both total order \textsc{Glucose}
implementations. The performance of this heuristic is likely due to
this large amount of deletions, coupled to the frequent conflicts
caused by partial order CDCL. Our intuition was that keeping as many
instantiations as possible would help building a model of the instance
faster, but apparently undoing as many instantiations as possible
is more useful. It indeed certainly allows to skip unsatisfiable parts
of the search space more quickly and to explore more various parts
of this space.

Heuristics based on counting instantiations to be undone during the
conflict have the drawback to be highly local, and consequently they
generally don't reach their goal globally. Indeed, the choice of the
assertion level doesn't only affect the current backtrack: dependencies
are added between this level and all other levels involved in the
conflict. These additional dependencies increase the likelihood for
the chosen assertion level to be deleted in future conflicts. If the
conflict clause involves $n$ decision levels (not including the conflict
level), the assertion level will have to depend on all other $n-1$
levels, but some of these dependencies may already exist. Intuitively,
picking the candidate assertion level which will entail the least
new dependencies should tend to globally lower the average quantity
of instantiations undone during a conflict. Conversely, we expect
the opposite heuristic to delete more instantiations per conflict.

For some unexplained reason, it is exactly the opposite that happens.
The \emph{least dependencies} strategy causes even more uninstantiations
than the \emph{most undos} heuristic, causing a slight increase of
solving time on unsatisfiable instances, but a further improvement
of performances on satisfiable instances. Figure~\ref{cactus-plot}
shows that with this heuristic 17 of the 20 satisfiable instances
are solved within 70 seconds, the 3 remaining instances being solved
in less than 500 seconds each. In contrast, 11 instances require more
than 100 seconds and 3 more than 1~500 seconds with the best performing
\textsc{Glucose} version.

On the other hand, the \emph{most dependencies} heuristic performs
poorly on satisfiable instances but very well on unsatisfiable instances.
Figure~\ref{cactus-plot} shows that it is by far the best tested
solver in terms of checked clauses and that it even steadily outperforms
the best \textsc{Glucose} version on all time limits.

This performance is explained by a sensible decrease of the average
number of undone instantiations per conflict compared to other PO-Glucose
implementations, as shown in Table~\ref{table3}. In the case of
unsatisfiable instances, undoing less instantiations seems to help
focussing the search on the currently active part of the search space.
Favorizing successive conflicts in related parts of the search space
results in a more efficient pruning and ultimately requires less conflicts
to prove unsatisfiability: regular \textsc{Glucose} with and without
phase saving need on average about 7 and 4,6 millions of conflicts
respectively for solving unsatisfiable benchmarks. This number drops
to between 2 and 2,7 millions of conflicts for previous \textsc{PO-Glucose}
variants, and down to 1,75 million with the \emph{most dependencies
}heuristic.

These dependencies-oriented heuristics and their contrasted efficiency
suggest that on SAT problems with low decision level interdependencies,
satisfiability solving can be significantly improved by using totally
different strategies depending on the actual satisfiability of the
instance: if a model exists, it can be found easier if backtracks
undo many instantiations, which helps exploring the search space more
dynamically. In the unsatisfiable case, backtracks should at the opposite
undo less instantiations to help focus the search on the currently
active search space and prove unsatisfiability with less conflicts.
Moreover, both types of strategies can be carried out by an appropriate
choice of assertion levels in a partial order CDCL search.

Satisfiability of instances with sparse level dependencies can thus
be very efficiently checked with PO-CDCL if the answer is known or
speculated prior to solving. We think it should be possible to design
a more balanced intermediate strategy that would perform significantly
better than total order CDCL regardless of the instance satisfiability.

\section{Conclusion}

In this paper, we further analyzed the partial order CDCL algorithm
and its behaviour on instances with sparse dependencies between decision
levels. We showed that the instantiations saved by the less destructive
backtrack of PO-CDCL often allow to discover conflicts dramatically
faster, which helps to prune the search space more efficiently. This
behaviour explains the good solving performances observed on tested
instances. Moreover, we noticed the significant impact of the assertion
level choice on the search and designed several heuristics for this
choice. According to our observations, opposite strategies are relevant
depending on whether the solved instance is or isn't satisfiable.
A satisfying model of the problem can be found faster if the backtrack
generally undoes large parts of the assignment, allowing quicker moves
in the search space. Conversely, undoing a smaller average quantity
of instantiations helps the solver to focus on the currently active
part of the search space and leads faster to a proof of unsatisfiability.
Finally, we showed that trying to locally control the amount of instantiations
undone by each individual backtrack is not the most efficient method;
heuristics that choose the assertion level according to the amount
of level dependencies it introduces have a stronger influence on the
average quantity of assignment deletions.

\bibliographystyle{splncs03}
\bibliography{bibliography}

\end{document}